\pdfoutput=1

\documentclass[11pt]{article}

\usepackage[]{emnlp2021}

\usepackage{times}
\usepackage{latexsym}
\usepackage{url}
\usepackage{mathdots}
\usepackage{yhmath}
\usepackage{cancel}
\usepackage{color}
\usepackage{siunitx}
\usepackage{array}
\usepackage{multirow}
\usepackage{amssymb}
\usepackage{tabularx}
\usepackage{booktabs}
\usepackage{svg}
\usepackage{tablefootnote}
\usepackage{cleveref}
\usepackage{tabularx}
\usepackage{xcolor}
\usepackage{makecell}
\usepackage{enumitem}
\usepackage{amsfonts}
\usepackage{pifont}
 \usepackage{subfigure}
\usepackage[T1]{fontenc}

\usepackage[utf8]{inputenc}

\usepackage{xspace}

\newcommand{\ours}{\mbox{C2F}\xspace}
\newcommand{\oursAblation}{C2F-NoHier\xspace}

\newcommand{\Ch}{f_\downarrow}
\newcommand{\Pa}{f^\uparrow}

\usepackage{microtype}

\newcommand{\smallsection}[1]{\vspace{1mm}\noindent{\textbf{#1.}}}

\newcommand\blfootnote[1]{%
  \begingroup
  \renewcommand\thefootnote{}\footnote{#1}%
  \addtocounter{footnote}{-1}%
  \endgroup
}

\title{Coarse2Fine: Fine-grained Text Classification on Coarsely-grained Annotated Data}

\author{
  Dheeraj Mekala$^1$  $\qquad$ Varun Gangal$^2$ $\qquad$  Jingbo Shang$^{1, 3, *}$ \\
  \small $^1$ Department of Computer Science and Engineering, University of California San Diego, CA, USA \\
  \small $^2$ Language Technologies Institute, Carnegie Mellon University, PA, USA\\
  \small $^3$ Hal\i c\i o\u glu Data Science Institute, University of California San Diego, CA, USA \\
  \small$^{1,3}$ \texttt{\{dmekala, jshang\}@ucsd.edu} $\qquad$ \small$^2$ \texttt{vgangal@cs.cmu.edu}
}

\begin{document}
\maketitle
\begin{abstract}
\blfootnote{$*$ Jingbo Shang is the corresponding author.}
Existing text classification methods mainly focus on a fixed label set, whereas many real-world applications require extending to new fine-grained classes as the number of samples per label increases. To accommodate such requirements, we introduce a new problem called coarse-to-fine grained classification, which aims to perform fine-grained classification on coarsely annotated data. 
Instead of asking for new fine-grained human annotations, we opt to leverage label surface names as the only human guidance and weave in rich pre-trained generative language models into the iterative weak supervision strategy. 
Specifically, we first propose a label-conditioned fine-tuning formulation to attune these generators for our task. 
Furthermore, we devise a regularization objective based on the coarse-fine label constraints derived from our problem setting, giving us even further improvements over the prior formulation. 
Our framework uses the fine-tuned generative models to sample pseudo-training data for training the classifier, and bootstraps on real unlabeled data for model refinement. 
Extensive experiments and case studies on two real-world datasets demonstrate superior performance over SOTA zero-shot classification baselines.
\end{abstract}

\section{Introduction}
\label{sec:intro}
\begin{figure}
    \centering
    \includegraphics[width=\columnwidth]{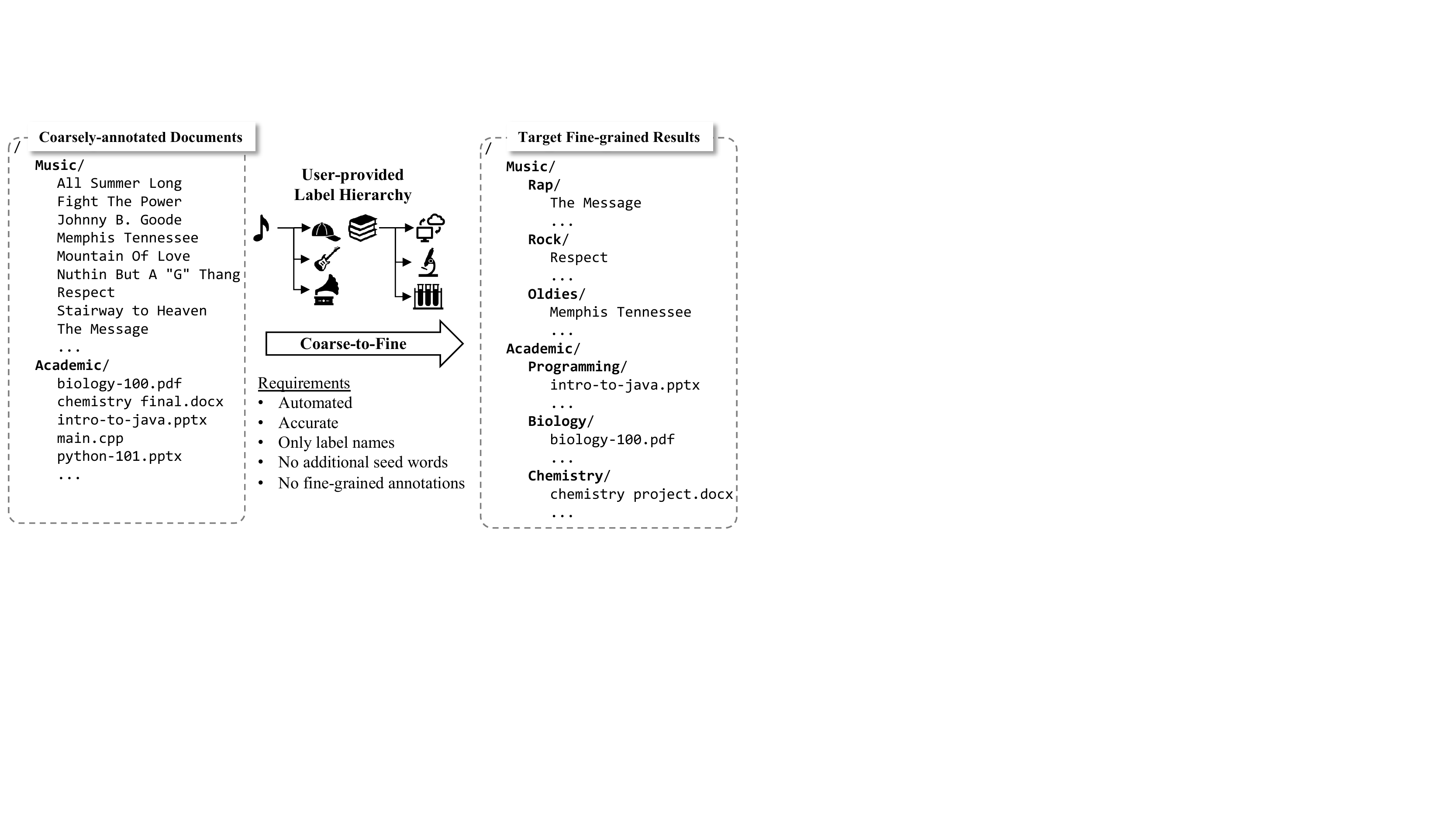}
    \caption{A visualization of our coarse-to-fine problem.}
    \label{fig:directory}
\end{figure}

In traditional text classification problems, the label set is typically assumed to be fixed.
However, in many real-world applications, new classes, especially more fine-grained ones will be introduced as the data volume increases. 
One commonly used method is to extend the existing label set to a label hierarchy by expanding every original coarse-grained class into a few new, fine-grained ones, and then assign a fine-grained label to each document. 
Using the directory structure for a set of files in computer as an example (see in Figure~\ref{fig:directory}), people usually start organizing the files in a coarse-grained fashion like ``Music" and ``Academics". 
Once the number of files in each of these coarse-grained directories increases, the categorization serves little purpose.
Therefore, we would like to create new fine-grained sub-directories inside coarse-grained directories like \{``rap'', ``rock'', ``oldies''\} for ``music" and similarly for ``academics''.
However, the process of assigning these files into fine-grained sub-directories typically begins with almost no supervision for fine-grained labels.

\begin{figure*}[t]
    \centering
    \includegraphics[width = \linewidth]{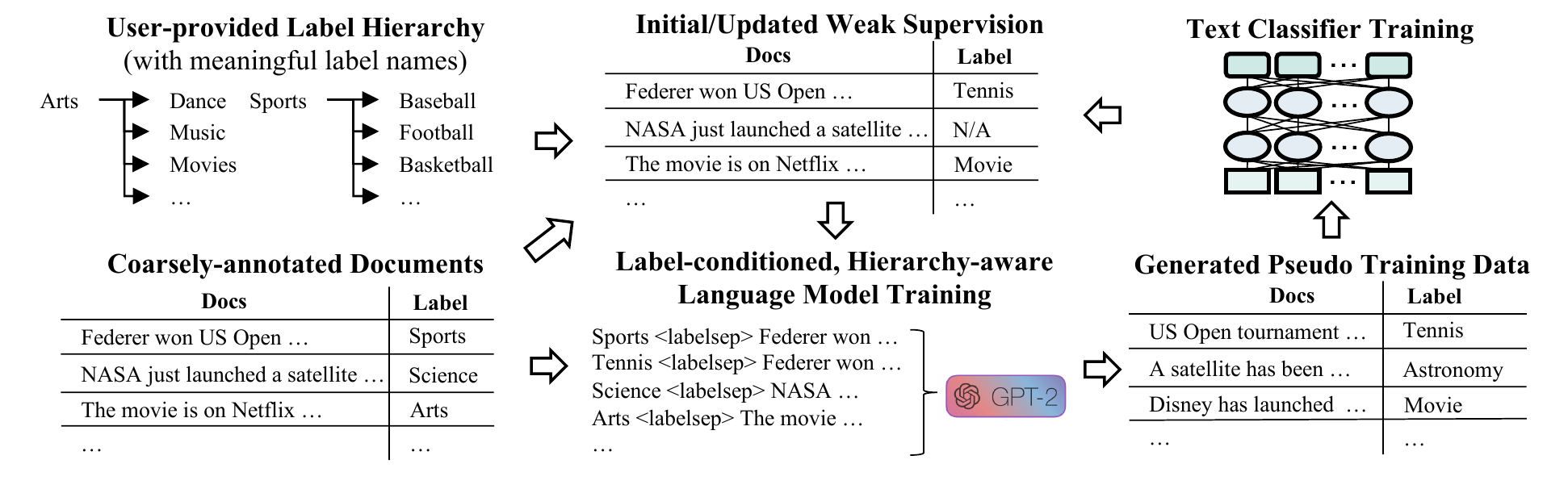}
    \vspace{-7mm}
    \caption{A visualization of our \ours framework. }
    \vspace{-3mm}
    \label{fig:overview}
\end{figure*}

To accommodate such requirements, in this paper, we introduce a new, important problem called coarse-to-fine grained classification, which aims to perform fine-grained classification on coarsely annotated data without any fine-grained human annotations.
There has been prior research on performing
text classification using extremely weak supervision, i.e., only label surface names as source of supervision.
For example, X-Class~\cite{wang2020x} learns class-aligned document representations to generate pseudo-labels and LOTClass~\cite{meng2020text} assumes replacements of label surface names in a sentence are related to the classes and leverages pre-trained language models to extract those words.
Note that, coarse-to-fine setting differs from generic zero-shot text classification in terms of having additional coarse supervision and a pre-conceived label hierarchy, though the final label set is available in either case.
And also, coarse-to-fine setting is different from hierarchical classification. We have no supervision for fine-grained labels other than the label names whereas the few shot hierarchical setting can have a few samples for fine-grained labels. 
Therefore, we want to capture the coarse-grained supervision and label hierarchy available to perform fine-grained classification.

In this paper, we propose a novel framework \ours as illustrated in Figure~\ref{fig:overview}. 
In the absence of fine-grained human annotations, it uses fine-grained label surface names as weak-supervision signals and leverages pre-trained language models as data generators. 
Similar to previous work, we first generate weak supervision from the whole corpus by assuming label surface names as their respective strong label-indicators.
For two iterations, \ours fine-tunes a language model based on weak supervision and trains a classifier based on generated pseudo-training data to refine weak supervision.
We observe that raw weak supervision usually has a highly-skewed label distribution, especially at the beginning, because the popularity of the label names varies. 
Since we have no prior knowledge about the underlying label distribution, to avoid significant deviations from that distribution, we opt to draw a balanced, weakly annotated subset through a stratified sampling before any model training.
We propose to fine-tune language models in a label-conditioned, hierarchy-aware manner.
Specifically, we inform the language models with label information by adding the label surface names at the beginning of each document. 
We further incorporate a regularization objective into the fine-tuning process that captures the constraints derived from the label hierarchy.
Facilitated by this fine-tuning process, we then generate pseudo-training data for each fine-grained label and train a classifier.
Next, using this fine-grained classifier's predictions over the coarsely annotated data, we select the samples with a high predicted probability for each respective fine-grained label.

We conduct experiments on two real-world datasets containing both coarse and fine-grained labels. 
The results demonstrate the effectiveness of our framework in leveraging a label hierarchy and a rich pre-trained language model to perform fine-grained text classification with no supervision.
Via thorough ablation, we isolate separate benefits accrued, initially just from using the label-conditioned, fine-tuned language model in the weak supervision pipeline, and the later incremental benefit once we incorporate our proposed regularization objective into the language model fine-tuning.

To the best of our knowledge, 
we are the first to work on the coarse-to-fine text grained classification, which aims to perform fine-grained classification on coarsely annotated data without any fine-grained annotations.
It is also worth mentioning that \ours is compatible with almost any generative language model and text classifier. 
Our contributions are summarized as follows:
\begin{itemize}[leftmargin=*,nosep]
    \item We develop a label-conditioned fine-tuning formulation for language models to facilitate conditional corpus generation.
    \item We devise a regularization objective based on the coarse-fine label constraints derived from the label hierarchy to be consistent with the pre-conceived label hierarchy.
    \item We conduct extensive experiments to demonstrate the superiority of \ours.
\end{itemize}
\noindent\textbf{Reproducibility.} We will release the code and datasets on Github\footnote{\url{https://github.com/dheeraj7596/C2F}}.
\subsection{Another Example Application}
\label{subsec:applicationExample}
Another example task which motivates how our framework could be deployed (besides the aforementioned directory example) is Intent Classification in Task-based Dialog \cite{chen2019bert,kumar2019closer,schuster2019cross,gangal2020likelihood}. This is often seen as a hierarchical classification problem \cite{gupta2018semantic}, with domains (e.g. movie, airline, shopping search) and intents (e.g. book-movie vs check-reviews, book-flight vs add-meal-options) forming the higher and lower levels of the label hierarchy. For a real-world task based dialog system (e.g. Alexa), there’s always a need over time to keep introducing both new domains (e.g. cruise, medical FAQ) and intents (order-popcorn, flight-entertainment) - as both data volume increases and the backend capabilities of the system expand.




\section{Problem Formulation}

The input of our problem contains:
(1) A tree-structured label hierarchy $\mathcal{T}$ with coarse-grained labels $\mathcal{C}$ at the first level and fine-grained labels $\mathcal{F}$ as their children. 
The $m$ coarse-grained classes are named \{$\mathcal{C}_1, \mathcal{C}_2, \ldots, \mathcal{C}_m$\},
and $k$ fine-grained classes are named as \{$\mathcal{F}_1, \mathcal{F}_2, \ldots, \mathcal{F}_k$\}.
All these class names are in natural language (e.g., words or phrases) and assumed to be informative;
and (2) a collection of $n$ text documents $\mathcal{D}$=\{$\mathcal{D}_1, \mathcal{D}_2, \ldots, \mathcal{D}_n$\} and their corresponding coarse-grained labels \{$c_1, c_2, \ldots, c_n$\}.


We record the mapping from each fine-grained class to its corresponding coarse-grained parent class as $\Pa$: $\mathcal{F}$ $\rightarrow$ $\mathcal{C}$.
The fine-grained classes in a coarse-grained label are represented by the coarse-to-fine mapping $\Ch$: $\mathcal{C}$ $\rightarrow$ $\mathcal{P}(\mathcal{F})$, where $\mathcal{P}(\cdot)$ is the powerset operator, which generates the set of all subsets. 
In this problem, each coarse class maps to a non-empty subset of fine classes, and all these subsets of fine-classes taken together are mutually non-overlapping and exhaustive.

We aim to build a high-quality document classifier from these inputs, assigning a fine-grained class label $\mathcal{F}_j \in \Ch(c_i)$ to each document $\mathcal{D}_i \in \mathcal{D}$.
\section{Our \ours Framework}

As visualized in Figure~\ref{fig:overview}, \ours aims to build a text classifier that can assign fine-grained labels to a set of coarsely-annotated documents based on only the label surface names and their hierarchical relations.
In the absence of fine-grained human annotations, it uses fine-grained label surface names as weak-supervision signals and leverages a pre-trained language model as data generators. 
Following an iterative process, \ours fine-tunes a language model based on weak supervision.
This fine-tuned language model is used to generate pseudo training data to train a fine-grained text classifier.
Based on the classifier's predictions, we select highly probable samples for each fine-grained class and repeat this process for one more iteration by replacing weak supervision with these samples.
This bootstrapping increases the quality of weak supervision by eliminating the mislabeled samples and improves the performance of text classifier as we show later in our case studies.

Our major contributions lie in how to better incorporate the label names and their hierarchical relations into the language model and therefore generate more high-quality psuedo training data.
Our framework is compatible with any generative language model and we choose GPT-2~\cite{radford2019language} in our implementation.
We feed label names to the language model through a label-conditioned formulation.
We further incorporate a regularization objective into the fine-tuning process that captures the constraints derived from the label hierarchy. The key components of \ours are discussed in detail in the following sections.

\subsection{Initial Fine-grained Weak Supervision}

We assume that user-provided label surface names are of high quality and are strong indicators for their respective classes, following the state-of-the-art weakly supervised text classification methods that only rely on label surface names~\cite{wang2020x,meng2020text}.
This assumption is intuitive and valid because there is no guidance other than class names from user and we expect them to be of high quality and indicative of the categories.

Ideally, the posterior probability of a document belonging to a class after observing the presence of strong indicators should be close to $1$.
Therefore, we consider samples that \emph{exclusively} contain the label surface name as its respective weak supervision.
Mathematically, let $W(\mathcal{F}_j)$ denote weak supervision of fine-grained class $\mathcal{F}_j$:
\begin{align*}
    W(\mathcal{F}_j) = \{\mathcal{D}_i | \mathcal{D}_i \cap \Ch(c_i) = \{\mathcal{F}_j\} \}
\end{align*}
where $\mathcal{D}_i \cap \Ch(c_i)$ returns a set of fine-grained label names under the coarse-grained class $c_i$ that appear in the document $\mathcal{D}_i$. 
When this set only contains $\mathcal{F}_j$, it means ``exclusive'' to other fine-grained labels. 
This ``exclusiveness'' could help us improve the precision of the initial weak supervision.
Note that, it is implied that $\mathcal{F}_j \in \Ch(c_i)$.

We observe that the initial weak supervision obtained usually has a highly-skewed label distribution, because the popularity of the label names varies.
This difference in distribution could bias the generative language model towards the majority label and might affect the quality of generated samples, which in turn, would affect the performance of the text classifier.
To address this problem, as there is no other prior knowledge, we opt to draw a balanced, weakly annotated subset through a stratified sampling before any model training.
In other words, we make the size of weak supervision uniform for all labels, equal to the size for the minority label.

\subsection{Tailored Language Model Training}

In this section, we describe our label-conditioned, hierarchy-aware language model training formulations that facilitates conditional corpus generation.
Specifically, we continuously train a pre-trained language model to capture the distribution $P(D|l)$, where $D$ is a document and $l$ is a (coarse or fine) label surface name.
Thus, this model can generate pseudo-training documents for fine-grained labels, when we plug in fine-grained label surface names.

\subsubsection{Label-Conditioned Generation}

Before we describe our formulation, we briefly introduce GPT-2 and its pre-training objective.  

\smallsection{GPT-2}
GPT-2 is a large, pre-trained left-to-right language model which exhibits strong performance with minimal in-task fine-tuning on many generation tasks, such as dialog~\cite{zhang2019dialogpt} and story generation~\cite{see2019massively}. 
Its strong zero-shot ability across tasks stems from its pre-training on the vast and diverse WebText corpus ($\approx$8M documents), besides the good inductive bias of its transformer-based architecture.
GPT-2 is trained on standard language modeling objective to maximize the likelihood of a document $D$ as follows:
\begin{align*}
    \mathcal{L}(D) = \sum_i \log P(w_i | w_{i-1}, \ldots, w_1; \Theta)
\end{align*}
where $P(\cdot)$ is modeled with a transformer-based architecture with parameters $\Theta$.

To continuously train GPT-2 in a label-conditioned way, one has to maximize $P(D|l)$ instead of $P(D)$.
We designate the label surface names as the special token sequences and append them in the beginning to their respective documents with another special token \texttt{<labelsep>} separating the label sequence and document.
For example, a sample document ``\emph{Messi plays for FC Barcelona}'' belonging to ``\emph{soccer}'' is modified to ``\emph{soccer \emph{\texttt{<labelsep>}} Messi plays for FC Barcelona}''.
Therefore, our objective is to maximize $\mathcal{L}(D | l)$ defined below:
\begin{align*}
    \mathcal{L}(D | l) = \sum_i \log P(w_i | w_{i-1}, \ldots, w_1; l; \Theta)
\end{align*}
Note that, the $l$ here could be the label surface name of a coarse-grained or fine-grained class.
One can view our formulation as asking the label token sequence to play the role of prompt and the document $D$ to be the continuation, thus facilitating conditional corpus generation.

During the continuous training process, we have access to both the gold, coarse-grained labels and weak, fine-grained labels.
Examples included in weak supervision give rise to two label-conditioned documents --- one by prefixing with the coarse-grained, gold label and the other with the weak, fine-grained one (due to the ``exclusiveness'' in the initial weak supervision).
Those not in the weakly supervised set only give rise to the first kind.
Since there is no conflict between these two labels, we simply treat a document as belonging independently to either of them.

\subsubsection{Hierarchy-Aware Regularization} 

Our label-conditioned generation treats both fine- and coarse-grained labels as prompts and does not use any information from the hierarchy.
Therefore, we propose to add a regularization to the language model with constraints derived from hierarchy.

Intuitively, fine-grained labels are more specific to coarse-grained labels, and therefore, when generating the same document conditioned on its gold fine-grained label, it should have a higher probability than that conditioned on its coarse-grained label.
We believe the same intuition is applicable to the high-quality weak supervision. 
Therefore, we seek to enforce the constraint while continuously training on weak supervision.
Specifically, a document should be more likely given its fine-grained (weak) label rather than its coarse-grained label.
Mathematically,
\begin{align*}
    P(\mathcal{D}_i | \mathcal{F}_j) > P(\mathcal{D}_i | c_i), \forall \mathcal{D}_i \in W(\mathcal{F}_j)
\end{align*}
where $W(\mathcal{F}_j)$ is the weak supervision for fine-grained label $\mathcal{F}_j$.
Note that, it is implied from $W(\mathcal{F}_j)$ that $\mathcal{F}_j \in \Ch(c_i)$.

This inequality can be expressed in the form of a margin between $P(\mathcal{D}_i | \mathcal{F}_j)$ and $P(\mathcal{D}_i | c_i)$, which can be implemented in practice through an additional Hinge loss term:
\begin{equation*}
      \small
      \begin{split}
          HL(\mathcal{D}_i, \mathcal{F}_j) = \max(0, \log P(\mathcal{D}_i|c_i) -\log P(\mathcal{D}_i | \mathcal{F}_j) + \epsilon)
      \end{split}
\end{equation*}
where $\epsilon$ is a positive constant. 

We incorporate this hierarchy-aware regularization into the final objective function as follows:
\begin{align*}
\begin{split}
    \mathcal{O} = & \sum_{\mathcal{D}_i \in \mathcal{D}} \mathcal{L}(D_i | c_i) \\
    & + \sum_{\mathcal{F}_j} \sum_{\mathcal{D}_i \in W(\mathcal{F}_j)} \mathcal{L}(\mathcal{D}_i | \mathcal{F}_j) - \lambda HL(\mathcal{D}_i, \mathcal{F}_j)
\end{split}
\end{align*}

The final optimization aims to maximize $\mathcal{O}$.


\begin{table*}[t]
    \center
    \vspace{-3mm}
    \caption{Dataset statistics}
    \label{tbl:datastats}
    \small
    \begin{tabular}{c c c c c c c}
        \toprule
            {\textbf{Dataset}} & {$|\mathcal{D}|$} & {$|\mathcal{C}|$} & {$|\mathcal{F}|$} & {\textbf{Coarse labels}} & {\textbf{Fine labels}} \\
        \midrule
        \textbf{NYT} & 11,744 & 5 & 26 & \begin{tabular}{@{}c@{}} \scriptsize arts, business, politics, \\ \scriptsize science, sports \end{tabular} &  \begin{tabular}{@{}c@{}} \scriptsize dance, music, movies, television, economy, energy companies, international \\ \scriptsize business, stocks \& bonds, abortion, federal budget, gay rights, gun control, \\ \scriptsize immigration,  law enforcement, military, surveillance, the affordable care act, \\ \scriptsize  cosmos, environment, baseball, basketball, football, golf, hockey, soccer, tennis\\ \end{tabular}\\
        \midrule
        \textbf{20News} & 16,468 & 5 & 17 & \scriptsize \begin{tabular}{@{}c@{}} \scriptsize computer, politics, recreation, \\ \scriptsize religion, science \end{tabular}&
        \begin{tabular}{@{}c@{}} \scriptsize graphics, windows, ibm, mac, x window, mideast, guns, autos, motorcycles, \\ \scriptsize baseball, hockey, christian, atheism, encryption, electronics, medicine, space  \end{tabular}
        
        \\
        \bottomrule
    \end{tabular}
\end{table*}

\subsection{Pseudo Training Data Generation, Text Classifier, \& Weak Supervision Update}
After continuously training the language model in a label-conditioned way, we generate the data for each fine-grained category.
Specifically, we send the corresponding label surface name as the prompt to our language model, and it then generates samples for that respective class. 
Since we don't know the label distribution beforehand, we assume it's a balanced distribution and thus, avoiding inducing potential bias in the classifier. 
We generate twice the required documents divided equally among fine-grained labels.
Specifically, for a fine-grained label $\mathcal{F}_j \in \Ch(c)$, we generate $2\frac{N_c}{|\Ch(c)|}$ documents, where $N_c$ is the number of documents that belong to coarse-grained label $c$.

We train a text classifier over these generated documents and their corresponding fine-grained labels.
Our framework is compatible with any text classifier and we use BERT(\verb+bert-base-uncased+)~\cite{devlin2018bert} classifier in our experiments.

After training the text classifier, we obtain fine-grained predictions and probability scores for all coarsely annotated documents $\mathcal{D}$.
Finally, we bootstrap it on unlabelled data by \emph{replacing} weak supervision $W(\mathcal{F}_j)$ by top-$k$ predictions where $k = |W(\mathcal{F}_j)|$ in every fine-grained label $\mathcal{F}_j$ and repeat this process \emph{one more time}.
In our experiments, we observe that these top-$|W(\mathcal{F}_j)|$ predictions are of significantly higher quality than the initial weak supervision, thus improving the text classifier. 

\section{Experiments}
In this section, we start with introducing datasets, compared methods, and experimental settings.
Next, we present quantitative evaluation results of \ours together with all compared methods.
In the end, we show qualitative studies to analyze different aspects of our \ours framework. 

\subsection{Datasets} 
We evaluate our framework on two hierarchical datasets where each document has one coarse-grained label and one fine-grained label. 
The dataset statistics are provided in Table~\ref{tbl:datastats}.
The details of these datasets are as follows:
\begin{itemize}[leftmargin=*,nosep]
    \item \textbf{The New York Times (NYT):} Following the previous work~\cite{meng2018weakly, mekala2020contextualized, wang2020x} we experiment on the NYT dataset. It is a collection of news articles written and published by The New York Times. Each news article is classified into one of 5 coarse-grained genres (e.g., arts, sports) and 25 fine-grained categories (e.g., movies, baseball).
    \item \textbf{The 20 Newsgroups (20News):} The 20News dataset\footnote{\url{http://qwone.com/~jason/20Newsgroups/}} is a collection of newsgroup documents partitioned widely into 6 groups (e.g., recreation, computers) and 20 fine-grained classes (e.g., graphics, windows, baseball, hockey). There are three miscellaneous labels (i.e., ``misc.forsale'', ``talk.politics.misc'', ``talk.religion.misc'').
    As one can notice, their label names are about `miscellaneous' and contain information of various types. 
    Since these labels and label surface names have no focused meaning, we drop the documents annotated as these labels in our experiments.
\end{itemize}

\subsection{Compared Methods}
Since we aim to perform fine-grained classification with no fine-grained supervision, we compare our framework with a wide range of zero-shot and weakly supervised text classification methods described below:
\begin{itemize}[leftmargin=*,nosep]
    \item \textbf{Word2Vec} learns word vector representations~\cite{mikolov2013efficient} for all words in the corpus and consider the word vectors of label surface name vectors as their respective label representations. In the case of multi-word label descriptors, the embeddings of individual words are averaged. Each document is labeled with the most similar label based on cosine similarity.
    \item \textbf{WeSTClass}~\cite{meng2018weakly} assumes words and documents share a joint semantic space and model each class as a high-dimensional spherical distribution. Words are sampled from this learned distribution to create pseudo-training data over which a classification model is trained. This model is refined through self-training on unlabeled documents. This is the same as applying its hierarchical counterpart WeSHClass~\cite{meng2019weakly} individually on each coarse-grained class.
    \item \textbf{ConWea}~\cite{mekala2020contextualized} is a seed-driven contextualized weak supervision framework. They leverage pre-trained language models to resolve interpretation of seed words and make the weak supervision contextualized.
    \item \textbf{LOTClass}~\cite{meng2020text} uses pre-trained language model like BERT~\cite{devlin2018bert} to query replacements for class names and constructs a category vocabulary for each class. This is further used to fine-tune the language model on a word-level category prediction task and identifies potential classes for documents via string matching. 
    A classifier is trained on this pseudo-labeled data with further self-training.
    \item \textbf{X-Class}~\cite{wang2020x} learns class-oriented document representations that make it adaptive to the user-specified classes. These document representations are aligned to classes through PCA + GMM, harvesting pseudo labels for a supervised classifier training.
\end{itemize}

We also compare \textbf{\ours} with its ablated variants.
\textbf{\ours-NoHier} uses label-conditioned generation alone without the hierarchy-aware regularization. \textbf{\ours-Ind} and \textbf{\ours-Ind-NoHier} are run individually on each coarse-grained label $c$ to assign a fine-grained label $\mathcal{F}_j \in \Ch(c)$ and the predictions are accumulated at the end to compute aggregated results. 
However, \textbf{\ours-Ind} uses both label-conditioned generation with the hierarchy-aware regularization whereas \textbf{\ours-Ind-NoHier} uses label-conditioned generation alone.
\textbf{\ours-1IT} is a BERT classifier trained on initial fine-grained weak supervision. 
We also consider \ours with different generative LMs and classifiers.
\textbf{\ours-GPT-BERT}, \textbf{\ours-GPT-LR} use GPT~\cite{radford2018improving} as the generative language model and BERT, Logistic Regression as classifiers respectively.

For a fair comparison, we make coarse-grained annotated data available for all baselines and run them individually on each coarse-grained label $c$ to assign a fine-grained label $\mathcal{F}_j \in \Ch(c)$ and the predictions are accumulated at the end to compute aggregated results. 
We provide label surface names as seed words for seed-word-driven baselines like ConWea and WeSTClass.

We also present the performance of BERT in a supervised setting which is denoted as \textbf{BERT-Sup}. 
The results of BERT-Sup reported are on the test set which follows an 80-10-10 train-dev-test split.

\subsection{Experimental Settings}

While fine-tuning GPT-2, we experiment with learning rates $\alpha \in \{5e^{-5},5e^{-4},5e^{-6}\}$, with $\alpha=5e^{-4}$ being found optimal, and continue the label-conditioned language model training for $5$ epochs. 

Generation from the model is done via nucleus sampling~\cite{holtzman2019curious}, with a budget of $p=0.95$ and a length limit of $200$ subwords. 
The prompt given for generation is simply the tag sequence corresponding to the intended fine-grained label of the sample to be generated. 
Since fine-grained class ratios are apriori unknown, an equal number of examples are sampled for each fine-grained class within the same coarse-grained class.

For the hierarchy-aware regularization, we set the hinge loss margin $\epsilon=\log 5$ and $\lambda=0.01$.
For hyperparameter selection of $\epsilon$, we sweep over the sequence of values in $\{log n\}_{n=1}^{n=10}$. Further searching is done through two levels of binary search. The decision to initially sweep over values in logarithmic fashion is taken based on two intuitions: i) Larger jumps were found to skip over the domain of variation of epsilon too quickly ii) $\epsilon$ is essentially a margin on logarithmic probabilities.

\begin{table}[t]
    \center
    \caption{Micro and Macro f1 scores and their respective standard deviations on two datasets are presented. The statistical significance test results between \ours and all other baselines are showed in Appendix \ref{subsec:statsig}. All the p-values are less than $10^{\text{-}15}$, making the performance improvement over baselines significant.}
    \vspace{-3mm}
    \label{tbl:f1_results}
    \small
    \setlength{\tabcolsep}{3pt}
\scalebox{0.8}{
    \begin{tabular}{l cc cc}
        \toprule
                         & \multicolumn{2}{c}{\textbf{NYT}} & \multicolumn{2}{c}{\textbf{20 Newsgroup}} \\
        \cmidrule{2-5}
        \textbf{Methods} & Mi-F$_1$(\%) & Ma-F$_1$(\%) & Mi-F$_1$(\%) & Ma-F$_1$(\%) \\
        \midrule
        Word2Vec & 32.50 (2.50) & 17.50 (1.50) & 11.03 (1.30) & 11.03 (0.90) \\
        ConWea & 76.23 (0.97) & 69.82 (0.54) & 56.14 (0.76) & 56.21 (0.32) \\
        WeSTClass & 73.96 (0.49) & 65.03 (0.31) & 55.46 (0.19) & 55.53 (0.38)  \\
        LOTClass & 15.00 (1.20) & 20.21 (0.76) & 34.18 (0.64) & 33.63 (0.71)  \\
        X-Class & 91.16 (0.56) & 81.09 (0.39) & 73.15 (0.23) & 73.06 (0.12) \\
        \midrule
        \ours & \textbf{92.62 (0.54)} & \textbf{87.01 (0.72)} & \textbf{77.50 (0.96)}  & \textbf{77.57 (0.89)}\\
        \oursAblation & 90.44 (0.91) & 85.50 (0.82) & 76.27 (0.85) & 76.13 (0.78) \\
        \ours-Ind & 91.60 (0.45) & 86.82 (0.44) & 74.62(0.96) & 74.50 (0.97) \\
        \ours-Ind-NoHier & 90.95 (0.59) & 85.75 (0.17) & 74.59 (0.63) & 74.48 (0.57) \\
        \ours-1IT & 82.10 (0.21) & 58.12 (0.31) & 71.06 (0.38) & 70.94 (0.53) \\
        \ours-GPT-BERT & 89.09 (0.34) &  85.07 (0.28) & 75.10 (0.64) & 75.02 (0.59) \\
        \ours-GPT-LR & 88.57 (0.48) & 84.98 (0.53) & 75.03 (0.63) & 74.89 (0.69) \\
        BERT-Sup  & 98.00 (0.27) & 94.00 (0.57) & 96.39 (0.43) & 96.36 (0.72) \\
        \bottomrule
    \end{tabular}
}
    \vspace{-3mm}
\end{table}

\subsection{Quantitative Results}
We evaluate our framework using Micro-f1(Mi-F$_1$) and Macro-f1(Ma-F$_1$) as performance metrics. 
The evaluation results of all methods run on three random seeds are summarized in Table~\ref{tbl:f1_results} along with their respective standard deviations.
We can observe that our proposed framework achieves superior performance compared to all other baselines.
We discuss the effectiveness of \ours as follows:

 
\begin{itemize}[leftmargin=*,nosep]
    \item \ours demonstrates the best performance among all compared baselines. By utilizing the generative language model through label-conditioned fine-tuning and regularizing it with hierarchical hinge loss to leverage the hierarchy, it is able to generate good quality pseudo training data, which helped in achieving the best performance.
    \item \ours outperforms X-Class with a significant margin. X-Class doesn't take advantage of label hierarchy and requires class names to be one word whereas our framework has no such limitation and leverages rich language models to understand informative label surface names.
    \item We have to note the significantly low performance of LOTClass. LOTClass queries replacements of label surface names and consider those to be indicative of the label. This is a valid assumption for the coarse-grained classification but when the classes become fine-grained, the replacements may not be indicative of its respective class. For e.g., consider the sentence ``I won a baseball game.''. If ``baseball'' is replaced by ``tennis'', it is still a valid and meaningful statement but ``tennis'' is not indicative of ``baseball''. Therefore, LOTClass performs low in the fine-grained text classification task. Our framework separates the weak supervision for each label initially and fine-tunes the language model in a label conditioned way. Therefore, it is able to distinguish between fine-grained labels as well.
    \item The comparison between \ours, \ours-Ind and \oursAblation, \ours-Ind-NoHier shows that the hinge loss helped in leveraging the constraints from hierarchy to improve the language model.
    \item The comparison between \ours and \ours-Ind shows that the fine-grained classification benefits from the hierarchical structure and joint training with other coarse-grained classes.
    \item We can observe that \ours perform significantly better than \ours-1IT. This shows that the fine-grained classification improves with bootstrapping, where the samples with high predicted probabilities are selected and used them as weak supervision for the next iteration.
    \item The comparison between \ours and \ours-GPT-BERT, \ours-GPT-LR shows that the performance improves with larger language models. This also demonstrates that \ours is compatible with different generative language models and classifiers.
    \item We observe that the performance of \ours is quite close to supervised method BERT-Sup, for e.g., on the NYT dataset. This demonstrates that \ours is quite effective in closing the performance gap between weakly supervised and supervised setting with just label surface names as supervision.
\end{itemize}

\subsection{Performance increase with bootstrapping}
\begin{figure}[t]
    \centering
    \includegraphics[width=\columnwidth]{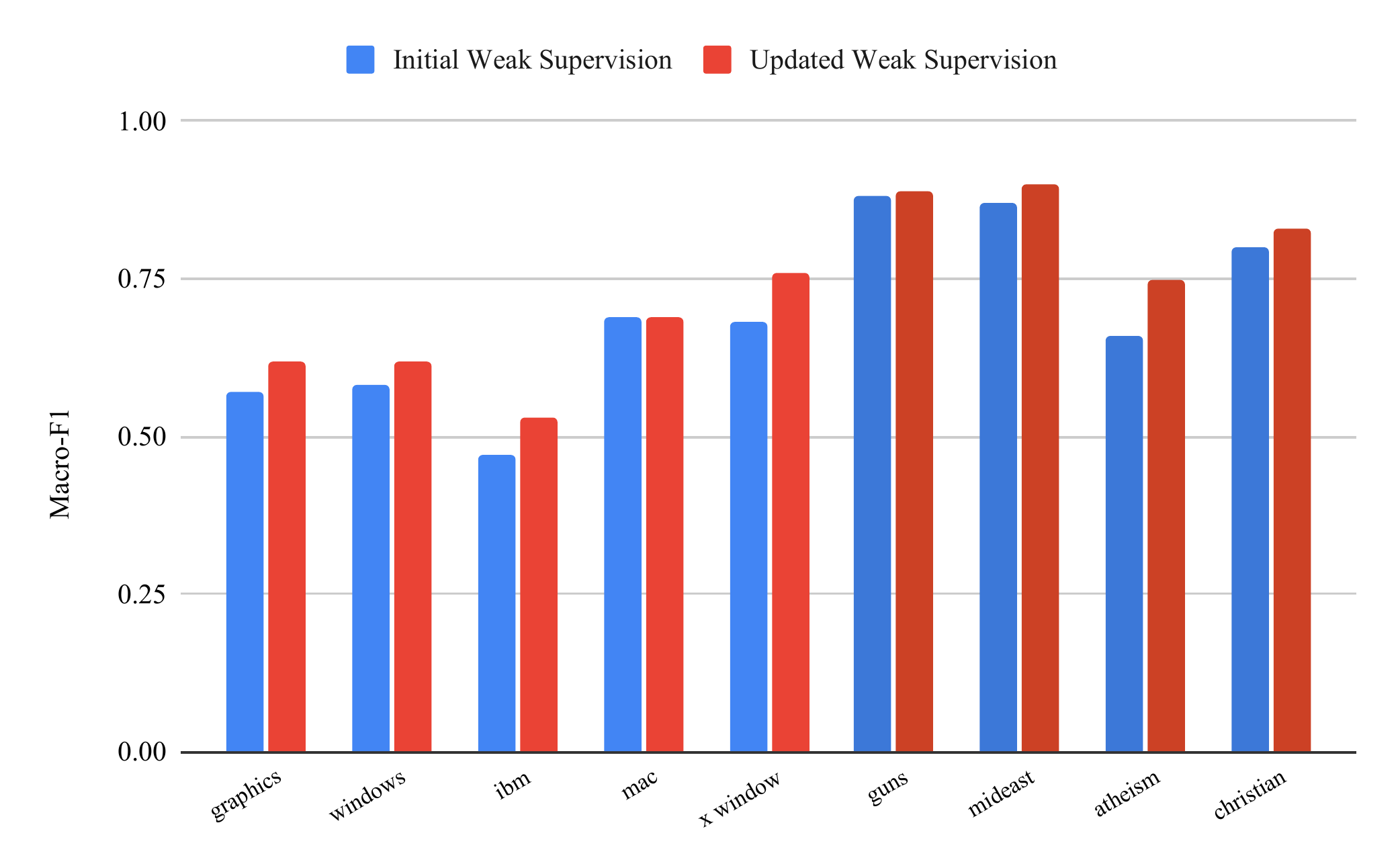}
    \caption{Performance increase in fine-grained text classifier from initial to updated weak supervision. 
    }
    \label{fig:relIt}
    \vspace{-3mm}
\end{figure}

The f1-scores of fine-grained labels in three coarse-grained labels ``computer'', ``politics'', ``religion'' across iteration-0 and iteration-1 are plotted in Fig~\ref{fig:relIt}. 
We see that performance increases significantly from iteration-0 (blue) to iteration-1 (red). 
We attribute this increase to our bootstrapping.

\subsection{Sensitivity to $\epsilon$}
\begin{figure}[t]
    \centering
    \includegraphics[scale=0.4]{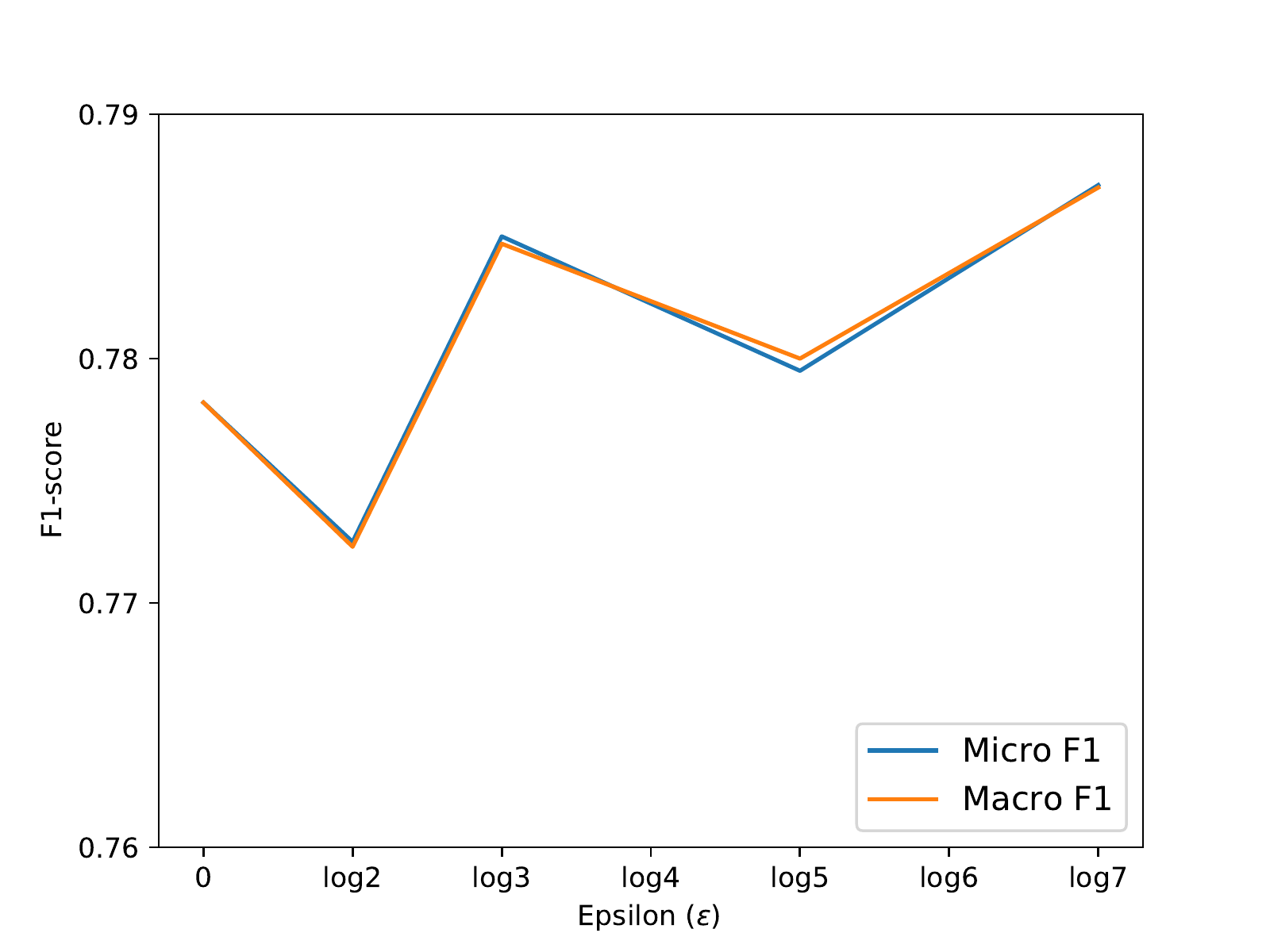}
    \caption{Micro and Macro-F1 scores vs $\epsilon$ on 20News. 
    }
    \label{fig:epsilon}
    \vspace{-3mm}
\end{figure}

A potential concern with the experimental setup can be overtly high sensitivity of C2F to the hinge loss margin parameter, i.e $\epsilon$. 
However, from the plot in Figure~\ref{fig:epsilon}, we clearly see that F1 scores aren't drastically sensitive to epsilon - with standard deviations of 0.00515 and 0.00517 for Macro and Micro-F1 scores respectively.

\subsection{Qualitative Analysis}
Given a particular coarse label (say \texttt{sports}) and its data subsets $X = \{\mathcal{D}_i|c_i=``sports"\}$ and $X_{f} = \{\mathcal{D}_i |{}c_i=``sports", f_i=f\in\Ch(``sports")\} $,  as a matter of post-hoc analysis, we can compare three distinct ``supervised" splits a classifier could've been trained on:
\begin{enumerate}[nosep,leftmargin=*]
    \item \textbf{Gold}: Data along with gold fine-grained labels, which is not actually available in our setting.
    \item \textbf{\ours-Init}: This is the subset of  $X$ for which the initial weak supervision strategy assigns fine labels based on label surface names.
    \item \textbf{\ours-Gen}: This is the data sampled from our trained language model as the generator for each of the respective fine-grained labels. 
\end{enumerate}
Which supervision is more apt from the purview of training? 
To answer this, we examine the entropy $H()$ 
of the word frequency distribution of the three datasets. 
Specifically, we examine reduction in value from entropy of the overall set $H(X)$ to the mean entropy on partitioning further by fine label, i.e $\bar{H}(X_{f})$. The larger this drop, the more internally coherent are the label partitions.

\begin{figure}[t]
    \centering
    \includegraphics[width=\columnwidth]{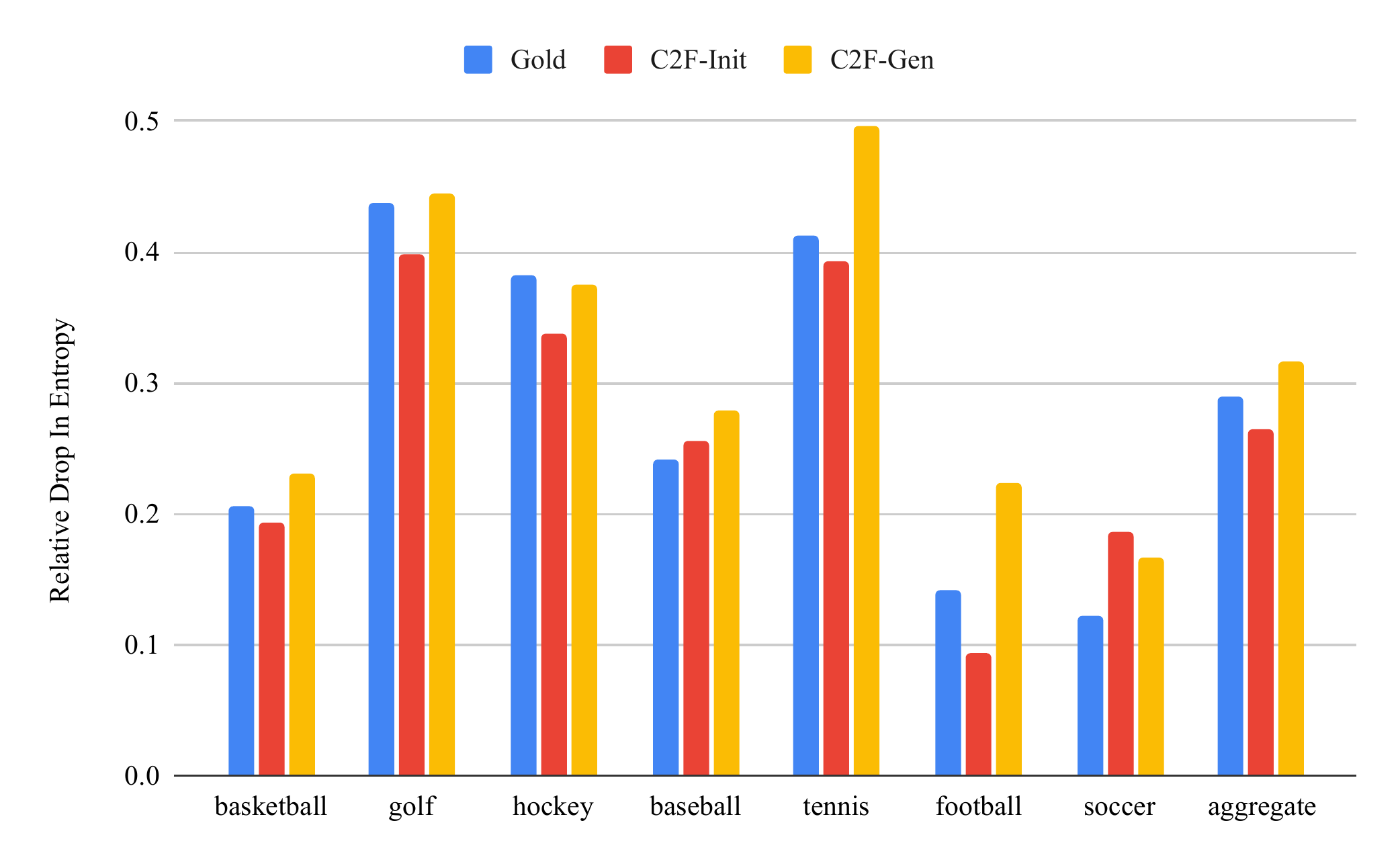}
    \caption{Relative drops in entropy $H(X)-H(X_{l_f})$ on splitting by fine label $l_f$, $\forall l_f \in  \Ch(\textsc{sports})$ , along with their aggregate. \textit{C2F-Gen, C2F-Init, Gold} stands for the generator sampled, initial weakly supervised subset and the entire ground truth datasets respectively.}
    \label{fig:relDropEntSports}
    \vspace{-3mm}
\end{figure}

As we can see from Figure~\ref{fig:relDropEntSports}, the drops $H(X)-\bar{H}(X_{f})$ are greater for \textsc{\ours-Gen} compared to both \textsc{\ours-Init} and \textsc{Gold}, indicating that it produces more mutually discriminative examples than both of them. At the same time, we observe that overall entropy of \textsc{\ours-Gen}, i.e $H(Gen)=6.631$ does not drastically differ in value from, though it is significantly lesser than,  that of \textsc{Gold}, $H(Gold)=6.924$, being 4.21\% smaller. In summary, we see that \textsc{\ours-Gen} provides a more discriminative training signal without reducing example diversity.  

A few samples of generated documents for fine-grained labels is shown in Table~\ref{tbl:Sample sentence}.

\begin{table*}[t]
    \center
    \caption{Example generated samples for fine-grained labels \textit{hockey}, \textit{basketball}, \textit{cosmos}, and \textit{economy} in NYT dataset and \textit{autos}, \textit{atheism}, \textit{windows} in 20Newsgroup dataset.}
    \label{tbl:Sample sentence}
    \vspace{-3mm}
    \small
    \begin{tabular}{c p{0.85\linewidth}}
        \toprule
            {\textbf{Label}} & {\textbf{Sentence}} \\
        \midrule
        hockey & \scriptsize the rangers' two injury-riddled offense and seven-game effort to defeat the florida panthers on saturday have prompted coach alain vigneault to call on the veteran right wing to start against the capitals. his message to the rookie goalie thomas vanek was that ``he was fine today'' and the rangers did not have to worry about his status in a game against another talented team. 
        \\
        \midrule
        basketball &  \scriptsize the nets entered the n.b.a. season with such wide open space that even their most ardent supporters of the franchise might admit their disappointment. the season ended sunday with the los angeles lakers selecting the first-round draft pick, with the eventual no. 1 pick jason kidd, a dynamic penetrator who could challenge for a starting job, at least in a year. 
        \\
        \midrule
        cosmos &  \scriptsize washington — the outermost reaches of space are barely perceptible from the surface, making it difficult to see the red planet, but a spacecraft based on this data could find something as yet elusive: an elusive galaxy that has just vanished.on thursday, kepler spacecraft, which is carrying back-to-back samples of solar starlight, will begin a journey that will lead the space agency\'s curiosity.\\
        \midrule
        economy & \scriptsize washington — the number of americans seeking unemployment benefits fell less than expected last week, the latest evidence that steady job growth is holding steady. jobless claims rose,000 to a seasonally adjusted,000, the labor department said on thursday. that is 146 more than previously reported.last week's drop in jobless claims came two months after steady increases in the previous week, when claims rose.\\
        \midrule
        autos & \scriptsize re manual shift bigots in article (james bruder) writes another question that hasn't been addressed yet is how come the auto mated to the lever controlling selector is not mounted to the transmission? i would think that the mated to the shifter would be mounted in the passenger compartment. is this a problem with the manual transmission? and if so, is it a problem with the shifter's mounting point?\\
        \midrule
        atheism & \scriptsize re why do people become atheists? in article, (kent sandvik) writes in article, (robert beauchaine) writes and i suppose i would have better evidence if i could. why would it be any different, for one thing? i\'m fairly new to this group, so perhaps this sort of question has already been asked, and answered before. but i\'ve just started to think about it.\\
        \midrule
        windows & \scriptsize re dos 6.0 in article 1qh61o, (russ sharp) wrote it's absolutely ludicrous for me to try and run dos 6.0 without the bloody help of at least 8 people. i've tried compiling it on several systems, and i've run it six times without a problem. dos 6.0 didn't mention a config.sys or anything else. there were a couple other windows' manuals which did mention about config.sys.\\
        \bottomrule
    \end{tabular}
\end{table*}


\section{Related Work}
We review the literature about different weakly supervised text classification methods.

There are three main sources of weak supervision: (1) a set of representative keywords for each class~\cite{meng2018weakly, mekala2020contextualized, mekala2020meta}, (2) a few labeled documents~\cite{tang2015pte, miyato2016adversarial, xu2017variational, meng2018weakly}, (3) label surface names~\cite{tao2018doc2cube, meng2020text, wang2020x}. 
Typically, weakly supervised text classification frameworks obtain pseudo-labeled data, train a classifier, and improve the classifier by bootstrapping over unlabeled data. 
Seed-driven frameworks obtain pseudo-labeled data from user-provided seed words. When a few labeled documents are provided as weak supervision, the above-mentioned pipeline similarly starts with these as pseudo-labeled data. 
In this paper, we focus on label surface names as the source of weak supervision. Along this line, Doc2Cube~\cite{tao2018doc2cube} expands label keywords from label surface names and performs multidimensional document classification by learning dimension-aware embedding; ~\cite{meng2020text} identify keywords for classes by querying replacements for class names using BERT and pseudo-labels the documents by string matching with the selected keywords. 
\cite{wang2020x} proposed an adaptive representation learning method for obtaining label and document embedding and these document embeddings are further clustered to pseudo-label the corpus.
However, all these methods perform flat text classification.
Although our method performs text classification using only fine-grained label surface names as supervision, we have coarse-grained annotated data and leverage it to improve fine-grained classification.
There are a few methods that perform weakly supervised hierarchical classification~\cite{meng2019weakly, zhang2021hierarchical}.
However, our problem statement is different from hierarchical classification.
We have coarse-grained annotated data and our framework utilizes it and label hierarchy to perform fine-grained text classification.
Recently, ~\cite{hsieh2019pseudo} introduced coarse-to-fine weakly-supervised multi-label learning problem. However, they assume a few fine-grained labeled documents as supervision whereas we require only label surface names.
Additionally, our framework is generative in nature i.e. instead of pseudo-labeling the corpus, we generate training data and train the classifier.
\section{Conclusions and Future Work}
Through this work, we introduced the task of coarse-to-fine grained classification and laid out its significance. Next, we showed the promise of incorporating pre-trained language models like GPT-2 into a weak supervision strategy which starts out with just label surface names. Finally, we showed a way to attune these models for our task even better, through explicit regularization based on coarse-fine label constraints which fall naturally out of our task definition. We outperform multiple SOTA zero-shot baselines on NYT and 20News, underscoring the utility both of incorporating pre-trained language models as well as task constraints.

We believe exploring newer ways of exploiting task agnostic knowledge sources and injecting task constraints into the weakly supervised learning process are promising avenues for future work.

\section{Acknowledgements}
We thank anonymous reviewers and program chairs for their valuable and insightful feedback. 
The research was sponsored in part by National Science Foundation Convergence Accelerator under award OIA-2040727 as well as generous gifts from Google, Adobe, and Teradata.
Any opinions, findings, and conclusions or recommendations expressed herein are those of the authors and should not be interpreted as necessarily representing the views, either expressed or implied, of the U.S. Government. 
The U.S. Government is authorized to reproduce and distribute reprints for government purposes not withstanding any copyright annotation hereon.


\bibliography{emnlp}
\bibliographystyle{acl_natbib}

\appendix
\clearpage
\newpage

\begin{Large}\textbf{Appendix}\end{Large}


\section{Stastical Significance Results}
\label{subsec:statsig}
We perform a paired t-test between \ours and each of the other baselines on both datasets and the results are showed in Table~\ref{tbl:t_test}. From these p-values, we can conclude that the performance improvement over baselines is significant.

\begin{table}[h]
    \center
    \caption{Statistical significance results.}
    \vspace{-3mm}
    \label{tbl:t_test}
    \small
\scalebox{0.95}{
    \begin{tabular}{c c c}
        \textbf{Baseline} & \textbf{p-value NYT} & \textbf{p-value 20News}\\
        \midrule
        ConWea & $8.34$ $\times$ $10^{-131}$ & $2.49$ $\times$ $10^{-165}$ \\
        WeSTClass & $5.18$ $\times$ $10^{-146}$ & $1.97$ $\times$ $10^{-166}$ \\
        X-Class & $6.45$ $\times$ $10^{-71}$ & $1.63$ $\times$ $10^{-92}$ \\
        \oursAblation & $1.80$ $\times$ $10^{-25}$ & $2.33$ $\times$ $10^{-55}$ \\
        \ours-Ind & $1.36$ $\times$ $10^{-18}$ & $7.92$ $\times$ $10^{-110}$\\
        \ours-Ind-NoHier & $3.46$ $\times$ $10^{-24}$ & $3.17$ $\times$ $10^{-114}$\\
        \bottomrule
    \end{tabular}
}
    \vspace{-3mm}
\end{table}

\end{document}